\documentclass{article}
\usepackage[title]{appendix}

\usepackage{spconf,amsmath,graphicx}
\usepackage{hyperref}
\usepackage{color}
\usepackage{float}
\usepackage{comment}
\usepackage{multicol}

\title{Where is the Fake? Patch-Wise Supervised GANs for Texture Inpainting}

\name{Ahmed Ben Saad$^{1,2,3}$, Youssef Tamaazousti$^{2}$, Josselin Kherroubi$^{2}$, Alexis He$^{3}$}
\address{
$^{1}$CMLA, ENS Paris-Saclay\\
$^{2}$Schlumberger AI Lab\\
$^{3}$Etudes et Productions Schlumberger \\
}

\begin{document}

\maketitle

\begin{abstract}

We tackle the problem of texture inpainting where the input images are textures with missing values along with masks that indicate the zones that should be generated. 
Many works have been done in image inpainting with the aim to achieve global and local consistency. 
But these works still suffer from limitations when dealing with textures. 
In fact, the local information in the image to be completed needs to be used in order to achieve local continuities and visually realistic texture inpainting. 
For this, we propose a new segmentor discriminator that performs a patch-wise real/fake classification and is supervised by input masks. 
During training, it aims to locate the generated parts (which we will call fake parts in consistency with the GAN framework), thus making the difference between real and fake patches given one image.
We tested our approach on the publicly available DTD dataset, as well as Electo-Magnetic borehole images dataset and showed that it achieves state-of-the-art performances and better  deals with local consistency than existing methods.
\end{abstract}
%

\section{Introduction}
\label{sec:intro}


The inpainting task consists in filling missing parts of an image. A "good" inpainting has to be visually plausible. In other words, it needs to respect the texture, colors, shapes and patterns continuities.
This is even more the case when we tackle \textit{Texture Inpainting}, which is the scope of this paper. 
Indeed, texture inpainting is of great interest in the oil and gas industry. 
For instance, borehole images\footnote{consists in "unwrapped" measures of resistivity (in Ohm/m) of the borehole, and is used to interpret geological layers of the subsurface}  usually contains missing data, due to the sensors design~\cite{gelman2017borehole,zhang2017structure}. 

\begin{figure}[tb!]
\label{vizabs}
\centering
\centerline{\includegraphics[width=8.5cm]{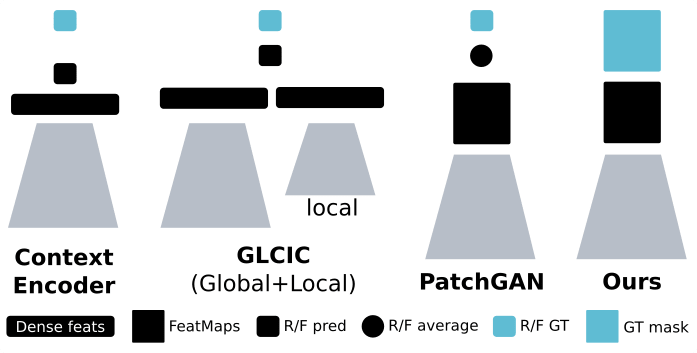}}
\vspace{-0.3cm}
\caption{
Visual comparison of s.o.t.a discriminators and our proposed one. CE : Each completed image is classified as Real (R) or Fake (F). GLCIC : A patch discriminator is added to classify a patch around the generated part as R/F. PatchGAN : The entire image is classified by aggregating classification scores of all the patches (mean score). Ours : The classification score of each patch is used and compared to the ground-truth (the inpainting mask).  
}
\vspace{-0.4cm}
\end{figure}

Generative Adversarial Networks \cite{goodfellow2014generative} proved to be very efficient in yielding the most realistic results in the inpainting task. 
For instance, Context Encoders (CE)~\cite{pathak2016context} (Fig.~\ref{vizabs} leftmost) obtained impressive results compared to traditional approaches~\cite{bugeau2009combining,criminisi2004region,drori2003fragment}. 
The idea was to train a generator (encoder-decoder network) with the help of an adversarial loss computed through a discriminator network. 
However, the main purpose of CE was feature learning and not inpainting, leading to a good global consistency (\textit{i.e.}, a generated image is globally visually plausible) but a poor local one (\textit{i.e.}, zooming on an image reveals many inconsistencies). 
 Iizuka \textit{et al.}, 2017 \cite{iizuka2017globally} tackled this problem of local inconsistencies, by adding a local discriminator (Fig.~\ref{vizabs} middle-left) that takes image patches centered on the completed region.  
 This technique succeeded in dealing better with local consistency but it usually generates boundary artifacts and distortions which forced the authors to use Poisson Blending \cite{perez2003poisson} as post-processing step. 
 Isola \textit{et al.}~\cite{isola2017image} went further by proposing a PatchGAN discriminator (Fig.~\ref{vizabs} middle-right) that divides the images into overlapping patches then classifies all of them. 
 The final output was the average of all classification results. 
 This technique was, for instance, successfully applied in inpainting in the medical imagery context by Armanious \textit{et al.}, 2018 \cite{armanious2019adversarial}. 
 However, we believe that averaging all the patches' contributions limits the power of the discriminators. 
 In fact, PatchGAN can classify images with tiny "fake" regions as globally real; and risk to learn features from the bad locations of fake and real regions. 
\begin{figure*}[tb!]
\centering
\centerline{\includegraphics[width=17.0cm]{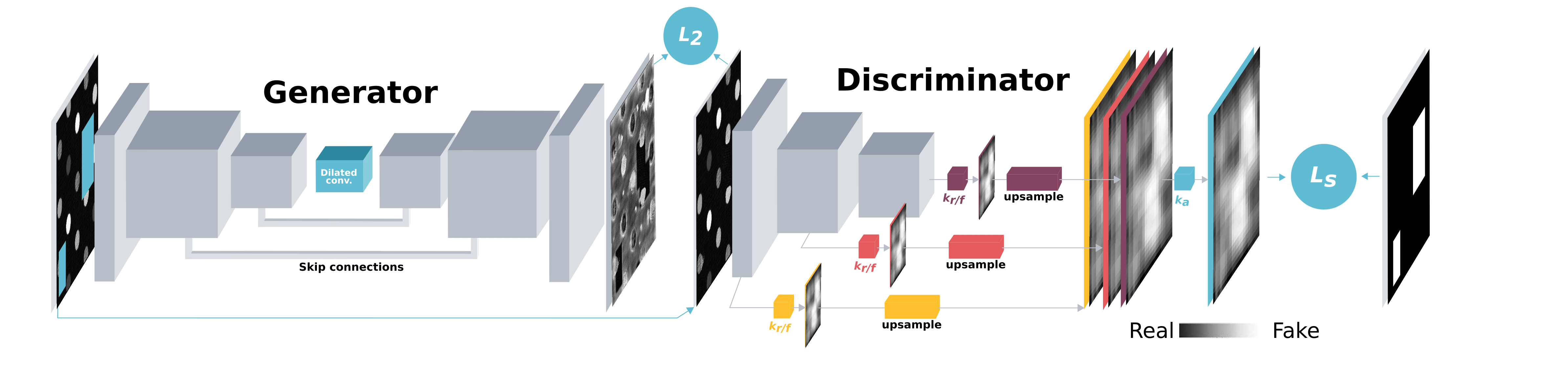}}
\vspace{-0.3cm}
\caption{
In our Inpainting framework, the generator (left) takes as input masked images and outputs inpainted images, that are fed to the discriminator (right) which segments fake regions. 
The latter is trained with GT masks and used as an adversarial loss for the former, which is also trained with classical reconstruction loss.
Feature maps are extracted at each scale (each color represents a scale), upsampled then concatenated before being fed the last 1x1 convolution (blue).
}
\vspace{-0.3cm}
\label{fig:method_description}
\end{figure*}
In this paper, we propose to solve these 
 problems using what we call a Segmentor As A Discriminator (SAAD). 
 The main idea behind SAAD (Fig.~\ref{vizabs} rightmost) is to have a finer discriminator that locates fake parts in inpainted images, thus backpropagates better gradients to the generator. To do so, instead of classifying the whole image as real or fake, we propose a discriminator that solves a segmentation task, and thus learn to locate the fake. 
 The segmentation ground-truth is given ``for free'' thanks to the inpainting masks.    
Additionally, while state-of-the-art (s.o.t.a) discriminators handle fake regions at one specific scale, we proposed to follow a multi-scale real/fake approach within our segmentor discriminator. 
Experiments were conducted on the DTD dataset~\cite{cimpoi14describing} where we compared our method to the works mentioned above. 
We also created a small dataset consisting of patches of borehole electro-magnetic images\footnote{Data provided by the Oil and Gas Authority and/or other third parties.} from 2 distinct wells and we used it for experiments. Details about this dataset are explained in Sec \ref{es}.
Results show that our approach achieves state-of-the-art performance and better inpaint texture and borehole images.

\section{Method Description}
\label{sec:pagestyle}

Our inpainting method is composed of two components: (i) a classical generator that performs the completion task (Sec.~\ref{gen}); and (ii) our main contribution that is a \textbf{S}egmentor \textbf{A}s \textbf{A} \textbf{D}iscriminator (SAAD) (Sec.~\ref{dis}). 

\subsection{Generator}
\label{gen}

The generator $\mathcal{G}$ takes as input masked images ($x_{mask} = x \odot (1-\mathcal{M})$ with $\mathcal{M}$ being the mask locations and $x$ the ground-truth image) and outputs inpainted images (denoted $\Tilde{x}_{final}$). 
$\mathcal{G}$ is a U-Net like architecture (encoder-decoder + long-skips)~\cite{ronneberger2015u} with 2-strided convolutions~\cite{springenberg2014striving} in the encoder-decoder for dimensions reduction and dilated convolutions~\cite{yu2015multi} in the middle convolutional blocks in order to increase receptive fields sizes. Details about the architecture are given as supplementary material.
We consider \textit{only} pixels at the mask locations. The final output of $\mathcal{G}$ is: $\Tilde{x}_{final} = x_{mask} +  \Tilde{x} \odot \mathcal{M}$.  
For the training of $\mathcal{G}$, we use the sum of a reconstruction loss $L_r$ as well as an adversarial loss $L_{adv}$ coming from our segmentor discriminator (described in the next section). 
For $L_r$, we use MSE between generated image and corresponding ground-truth (GT): $L_{r}(x,\Tilde{x}_{final},\mathcal{M})~=~\| x\odot \mathcal{M} - \Tilde{x}_{final} \odot \mathcal{M}\|_{2}^2$. 

\subsection{Segmentor As A Discriminator (SAAD)}
\label{dis}
The main idea behind SAAD is to have a finer discriminator that is able, given an inpainted image, to locate its fake parts, thus backpropagating better gradients to the generator. 
Locating the fake helps in: (i) avoiding to classify images with tiny generated regions as globally real or fake; and (ii) learning features from the correct locations of fake and real regions. 
To locate the fake, we propose that the discriminator performs a segmentation task. 
In fact, in inpainting, the segmentation masks are given ``for free'', since they correspond to the inpainting masks.  
Specifically, the discriminator $\mathcal{D}$ takes as input $x_{final}$ and outputs feature maps $\mathcal{F}_{feats}$ on top of which we add a convolution filter $k_{r/f}$ that outputs a real/fake map that we denote $\mathcal{F}_{r/f}$. 
Simply said, $\mathcal{F}_{r/f} = k_{r/f} (\mathcal{F}_{feats})$. 
To learn our segmentor discriminator $\mathcal{D}_S$, we enforce its output $\sigma(\mathcal{F}_{r/f})$ ($\sigma$: sigmoid function) to be close to $M$, by minimizing a pixel-wise BCE loss. This corresponds to $L_{s}$. 
Note that, for $\mathcal{D}$ we can use classical architectures, thus, the output size of the last feature map is usually \textit{smaller} than the input size. 
It is thus the same for $\mathcal{F}_{r/f}$. 
Hence, to match the size of the input masks ($h$$\times$$w$), we up-sample $\mathcal{F}_{r/f}$ from $h'$$\times$$w'$ to $h$$\times$$w$. 
Note also that $k_{r/f}$ has a receptive field of size $s$$\times$$s$ with $s$$>$$1$. 
This means that $k_{r/f}$ classifies \textit{patches} of the input images and this is why we characterize $\mathcal{D}_S$ as a \textit{patch-wise} discriminator. 
In order to handle real/fake patches at different scales we propose to follow a multi-scale real/fake segmentation approach. This is essential for capturing more texture diversity. 
For this, we perform the segmentation task with multiple filters positioned at different levels of the network and thus having different receptive fields sizes. 
Formally, each filter $k_{r/f}^i$ takes as input the feature maps given by the $i^{th}$ convolutional layer and outputs real/fake maps $\mathcal{F}_{r/f}^i$ that are upsampled, concatenated and fed to the last 1 $\times$ 1 $\times$ 1 convolutional filter $k_a$ to generate the segmentation map. The latter is compared to the ground-truth mask $\mathcal{M}$. 



\begin{table*}[tb!]
\centerline{}
\centering
\centering
\begin{tabular}{l|lll|lll}
\hline
\textbf{Method} & \multicolumn{3}{c|}{\textbf{DTD}} & \multicolumn{3}{c}{\textbf{Boreholes}} \\ \hline
 & MPS & PSNR & SSIM & MPS & PSNR & SSIM \\
\textbf{CE} (Pathak et al. \cite{pathak2016context}) & 95.3 & 24.385 & 0.901 & 91.0 & 24.890 & 0.878 \\
\textbf{GLCIC} (Iizuka et al. \cite{iizuka2017globally}) & 96.2 & 24.728 & 0.924 & 92.4 & 24.944 & 0.889 \\
\textbf{GLPG} (Isola et al. \cite{isola2017image}) & 95.6 & 26.409 & 0.930 & 95.0 & 26.152 & 0.919 \\
\hline
\textbf{SAAD} (ours) & \textbf{97.3} & \textbf{27.536} & \textbf{0.937} & \textbf{96.3} & \textbf{27.085} & \textbf{0.931} \\ \hline
\end{tabular}
\caption{Comparison of our method to the state-of-the-art. PSNR in dB and MPS in \%}
\label{tab:Table1}
\end{table*}

\section{Experiments and results}
\label{sec:typestyle}
\begin{figure}
\includegraphics[width=8.5cm]{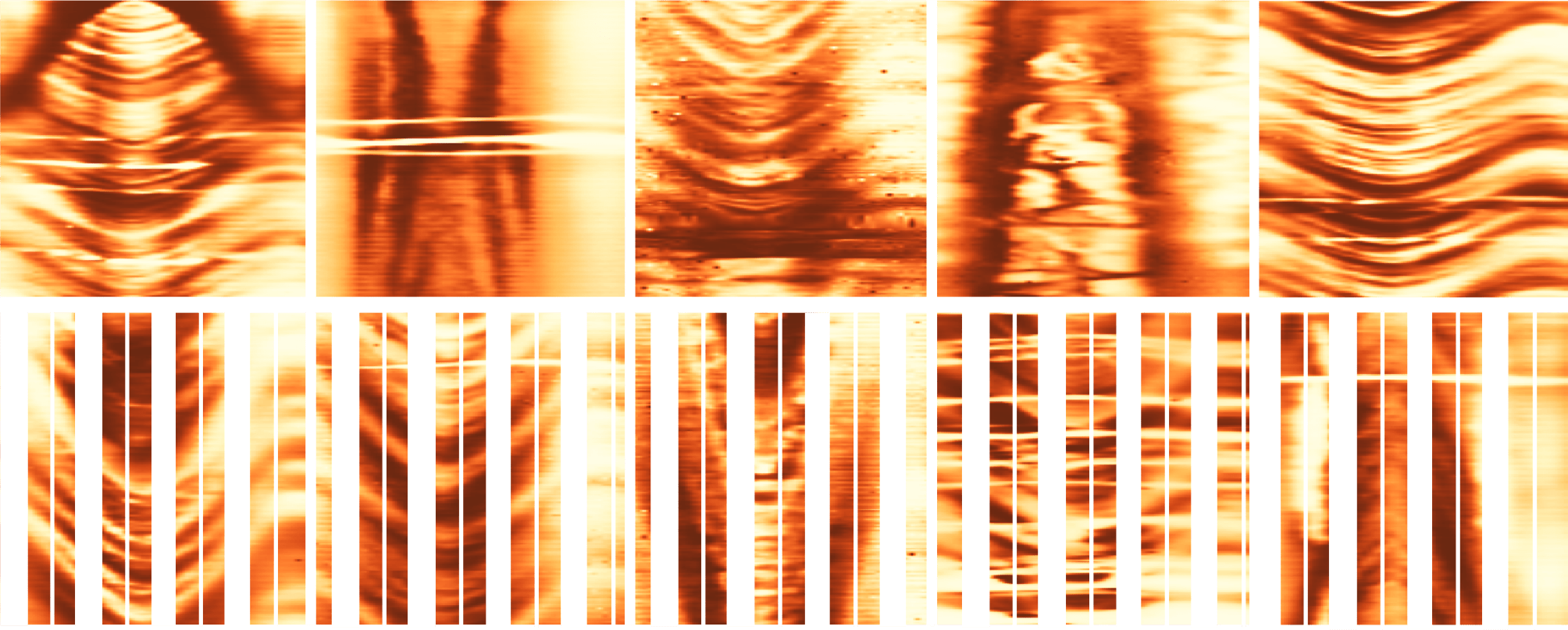}
\vspace{-0.3cm}
\caption{Examples of boreholes images with each column corresponding to the azimuth of the borehole and each row to the depth within the well. 
On top, we show images from a full horizontal-axis coverage sensor and at bottom, we show images from a sensor that has missing values. 
}
\vspace{-0.1cm}
\label{examples}
\end{figure}


\subsection{Experimental settings}
\label{es}
\noindent \textbf{Texture Inpainting Task}\\
Since the GAN-based Texture Inpainting task is not common in the literature, we proposed to set up a new experimental protocol using, first the publicly available Describable Textures Dataset (DTD)~\cite{cimpoi14describing} as well as our collection of borehole images dataset. 
The former (DTD) contains 5640 texture images and we used nearly 200 random images for testing purposes and the rest for training/validation. 
For each image, we generated multiple rectangle masks (random number, at most 5), at randomly positions before feeding it to the generator. 
The masks eventually overlaps each other and cover 15\% to 30\% of training and test images.

\begin{figure}[tb!]
\includegraphics[width=8.0cm, height=6cm]{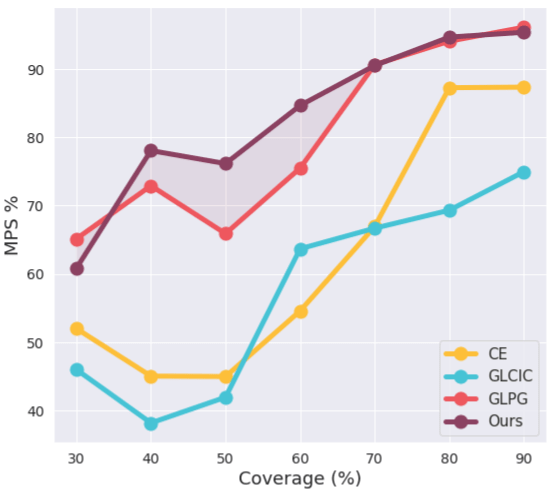}
\vspace{-0.3cm}
\caption{Models performances (MPS) with respect to coverage proportion on boreholes dataset. Results for PSNR and SSIM are given in the supplementary material.}
\label{coverage}
\end{figure}

For the borehole dataset -- that consist in "unwrapped" measures of resistivity (in Ohm/m) of the borehole, and is used in the oil and gas sector to interpret geological layers obtained while drilling wells --, we collected released data from the UK NDR website\footnote{\url{https://ndr.ogauthority.co.uk}}, and masked it following the common masks that we have in drilling tools (\textit{i.e.}, vertical rectangular gaps with a regular spacing due to the tool design). 
Specifically, we roughly used $2500$ 256$\times$256 overlapping patches of the resulting borehole data taken from \textit{one} well and the test set consist of nearly 2000 patches taken from a different well. 
This dataset will be released and some example images are given in Fig~\ref{examples}.

For both datasets, we used a fixed set of masks for the test images for fair comparisons. 
Indeed, to compare the performances of all methods, we used 3 common metrics: Peak Signal To Noise Ratio (PSNR), Structural Similarity (SSIM) and Mean Perceptual Similarity (MPS) computed by: $\frac{1}{card(X)} \sum_{x \in X} (1~-~PS(x,\Tilde{x}))$, where $X$ is the set of masked test images, and PS is the Perceptual Loss as defined in~\cite{zhang2018unreasonable}. 
Moreover, for statistical accuracy reasons, we trained every method $5$ times and report the average score.

 \noindent \textbf{Comparison Methods}\\
We compared our discriminators with three existing approaches: (i) Context Encoder (CE) that globally classifies the generated image; (ii) GLCIC which consist in concatenating the features of a global and a local discriminator; and (iii) GLPG which is a combination of GLCIC and PatchGAN (consist in classifying real/fake patches with convolutional filters and averaging their outputs to get the global prediction). SAAD and these three methods are illustrated in Fig~\ref{vizabs}. 
One should note that, many works proposed to use Perceptual loss~\cite{zhang2018unreasonable,armanious2019adversarial} calculated over pretrained features~\cite{tamaazousti2019learning} but this is orthogonal to our contribution, and the goal here is to asses the different supervisions of the discriminators.  
Note that, the same generator network was used for all the methods as well as the same discriminator's backbone. The latter, corresponding to the first 3 blocks of the ResNET-18~\cite{he2016deep} architecture as we are dealing with textures and do not need high-level features (see supplementary material for more details). For the local discriminator in GLCIC and GLPG, we used just the two first blocks. 
We trained all the networks with 200 epochs using Adam optimizer with learning rates of $10^{-4}$ and $4\times10^{-4}$ respectively for the generator and the discriminator. 
To avoid model collapse, we used zero-centered gradient penalty as defined in~\cite{mescheder2018training}.



\begin{figure}[tb!]
\includegraphics[width=8.5cm]{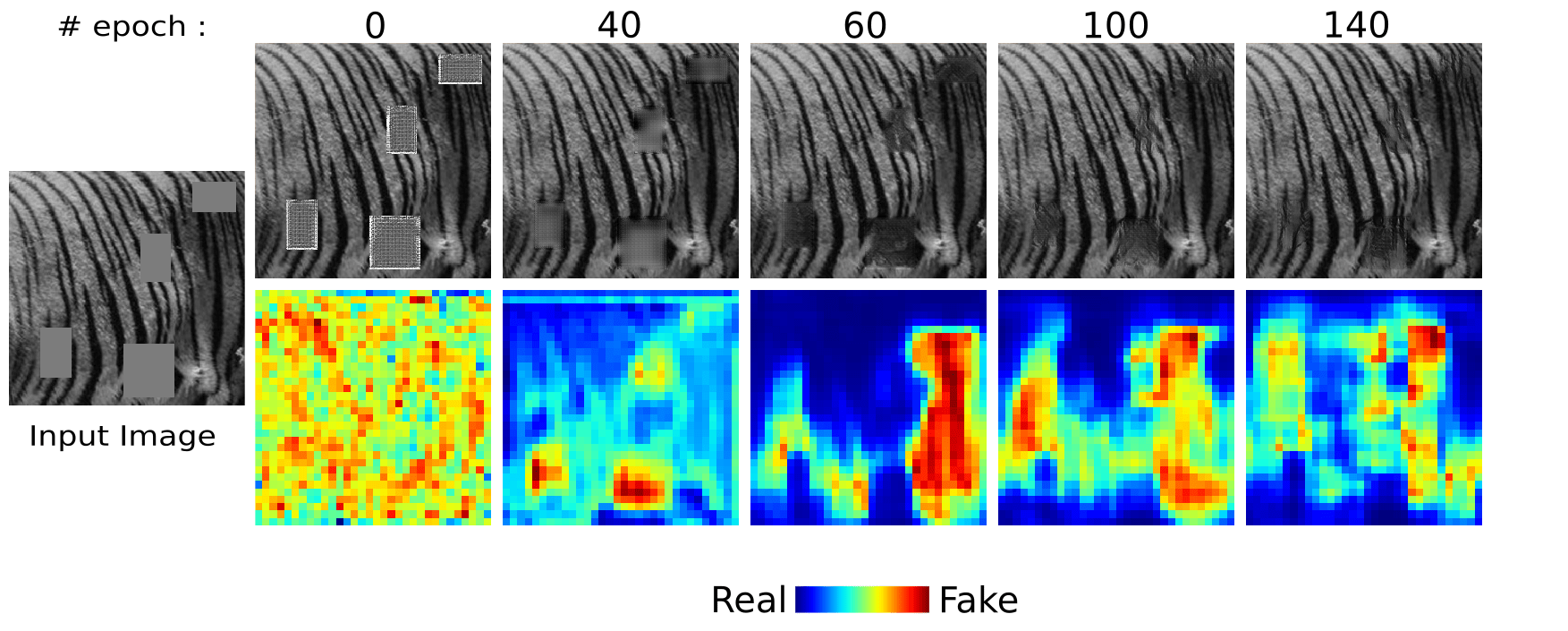}
\vspace{-0.3cm}
\caption{Evolution of the segmentation map for a DTD image. \textbf{Top} : Inpainted images. \textbf{Bottom} : Segmentation maps.}
\label{evo}
\end{figure}

\begin{figure}[tb!]
\includegraphics[width=8.5cm]{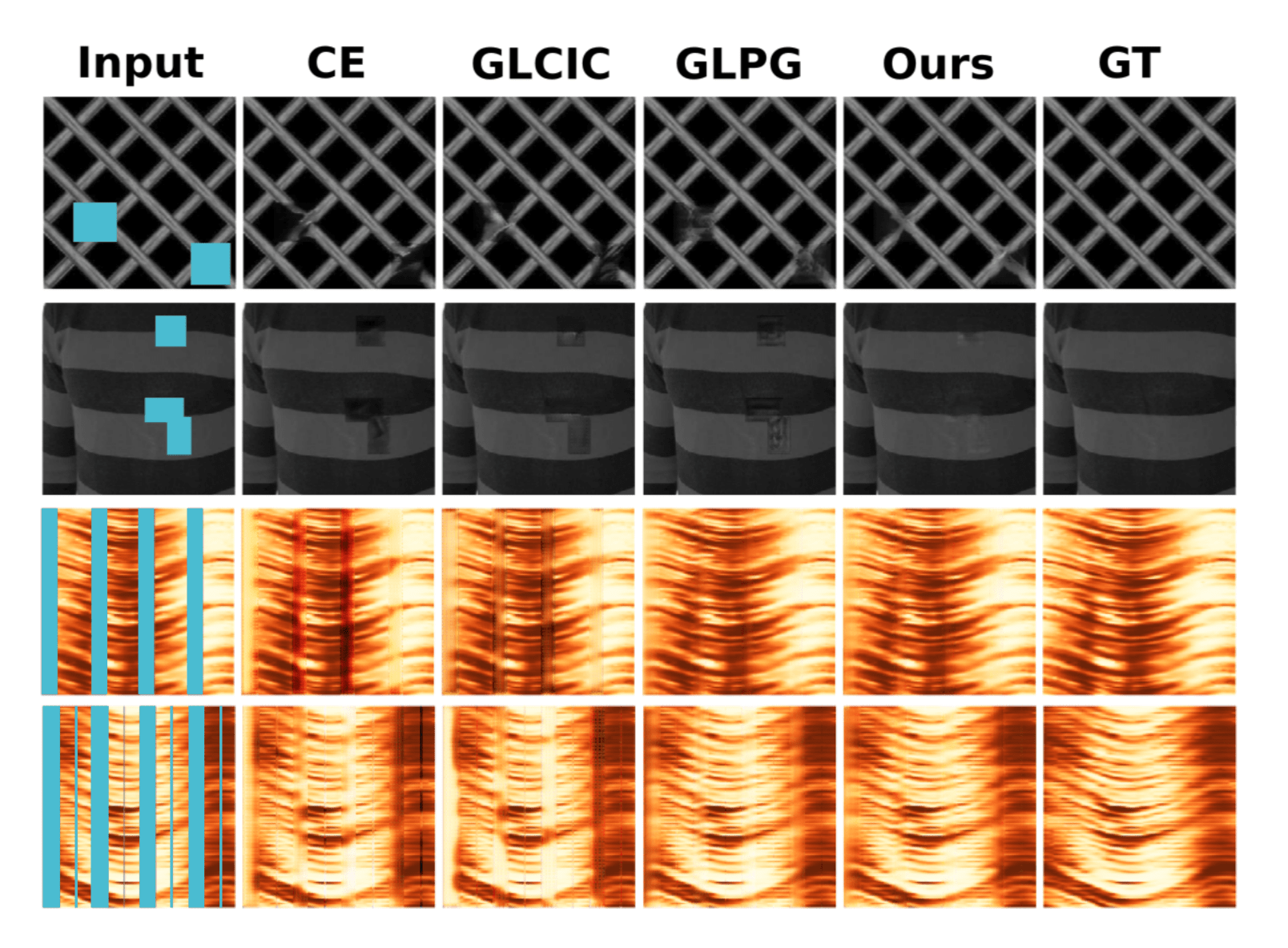}
\vspace{-0.3cm}
\caption{Qualitative results of different methods on DTD and borehole dataset. \textbf{Top} : DTD images. \textbf{Bottom} : Borehole images. Masks are colored in blue for visibility.}
\label{qlttve}
\end{figure}

\vspace{-0.4cm}
\subsection{Results}
The results of the different methods on the texture inpainting task on DTD and on borehole dataset are presented in Tab.~\ref{tab:Table1}.
We can see that our methods perform better than all others, regardless of the evaluation metric. 
For instance, on DTD, SAAD outperforms the CE baseline by 2 points of MPS. The difference grows up to 5 point on the borehole dataset.
In addition, compared to the recent GLPG, we improve the MPS by 1.7\% on DTD and by 1.3\% on borehole dataset. 
Since the only difference between GLPG and SAAD is the supervision (\textit{i.e.}, classification vs segmentation), this result shows that the main contribution of this paper is valuable. 
Some qualitative results are shown in Fig.\ref{qlttve}. Showing better inpainting done by our method compared to existing methods. More qualitative results are given as supplementary material.

\begin{figure}[tb!]
\centering
\includegraphics[width=8.5cm]{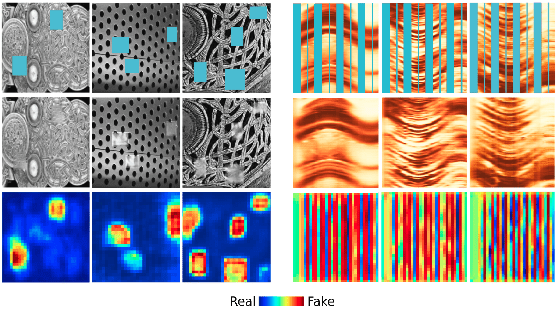}
\vspace{-0.3cm}
\caption{Segmentation maps of our discriminator at random epoch during training. \textbf{Top} : Input images. \textbf{Middle} : Inpainting results. \textbf{Bottom} : Discriminator's segmentation maps.}
\label{fm}
\end{figure}
\vspace{0.2cm}

\noindent{\textbf{Performance with respect to coverage}}  \\
We studied the effect of coverage ratio (low/high coverage : large/thin masks) on the models' test performances. For this we fixed the size of the masks at each step and put them at random positions on the training images. We then trained each method on the borehole dataset. The test results are shown in Fig.\ref{coverage}. We can see that our method performs better on low and medium coverages (especially up to 60\%) where the difference in PSNR, MPS and SSIM is visible ($+ 0.1$ in SSIM, $+ 0.7\%$ in MPS between GLPG and SAAD for $40\%$ coverage) 
As the coverage increases, the inpainting task becomes much easier which explains the small difference in performances between methods. We conclude that our method is well suited for inpainting tasks with relatively large zones to inpaint.

\vspace{0.2cm}

\noindent{\textbf{Segmentation Maps visualization}} \\
To exhibit the behaviour of SAAD during training, we visualized the segmentation maps given by the discriminator during training (before convergence) on different examples from both datasets. We see on Fig.\ref{fm} examples where, in the middle of the training process, the discriminator becomes able to roughly \textit{locate the fake zones} in the images that are generated by the generator network. And sometimes misclassifies some patches (Fig.\ref{fm}). This drives the generator updates towards yielding more realistic inpaintings.
Visual examples of evolution of the segmentation map during training are given in Fig.\ref{evo}. This confirms our intuition about the discriminator behaviour in the middle of the training process and shows how at the end (epoch 140), the segmentation task becomes more and more difficult for SAAD. Which means that our model is converging.



\section{Conclusion}
We presented a new approach for GAN-based texture inpainting that involves changing the discrimination task to a segmentation one to achieve better texture completion. 
We have shown, through quantitative and qualitative results on DTD and on a borehole images dataset, that this new way of supervision allows the generator to better generate textures and preserve mostly local features like colors, contrasts and shapes.



\small

\bibliographystyle{IEEEbib.bst}
\bibliography{main}

\end{document}


\onecolumn

\maketitle

\section{Appendix}
The supplementary material provides a detailed overview of the SAAD architecture for reproducibility reasons, extra qualitative comparisons of our method with existing baselines on DTD and boreholes dataset, as well as a comparison of models performances with respect to coverage for differents methods in PSNR and SSIM.
\section{Details of architectures}
\begin{figure*}[h!]
\center
\centerline{\includegraphics[width=17cm,height=10cm]{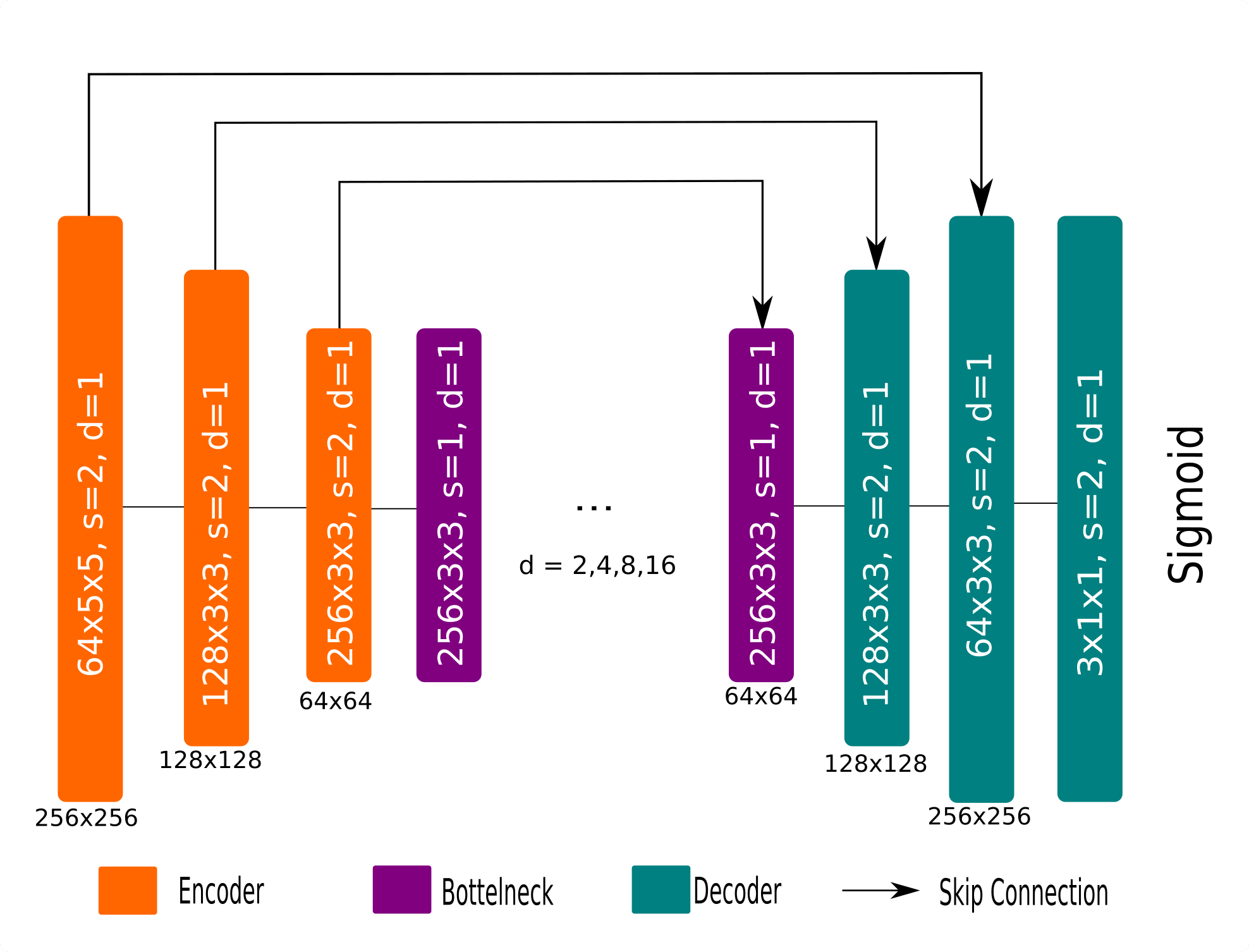}}
\vspace{-0.3cm}
\caption{Detailed overview of the Generator.}
\label{gen}
\end{figure*}

\begin{figure*}[h!]
\center
\centerline{\includegraphics[width=17cm,height=9cm]{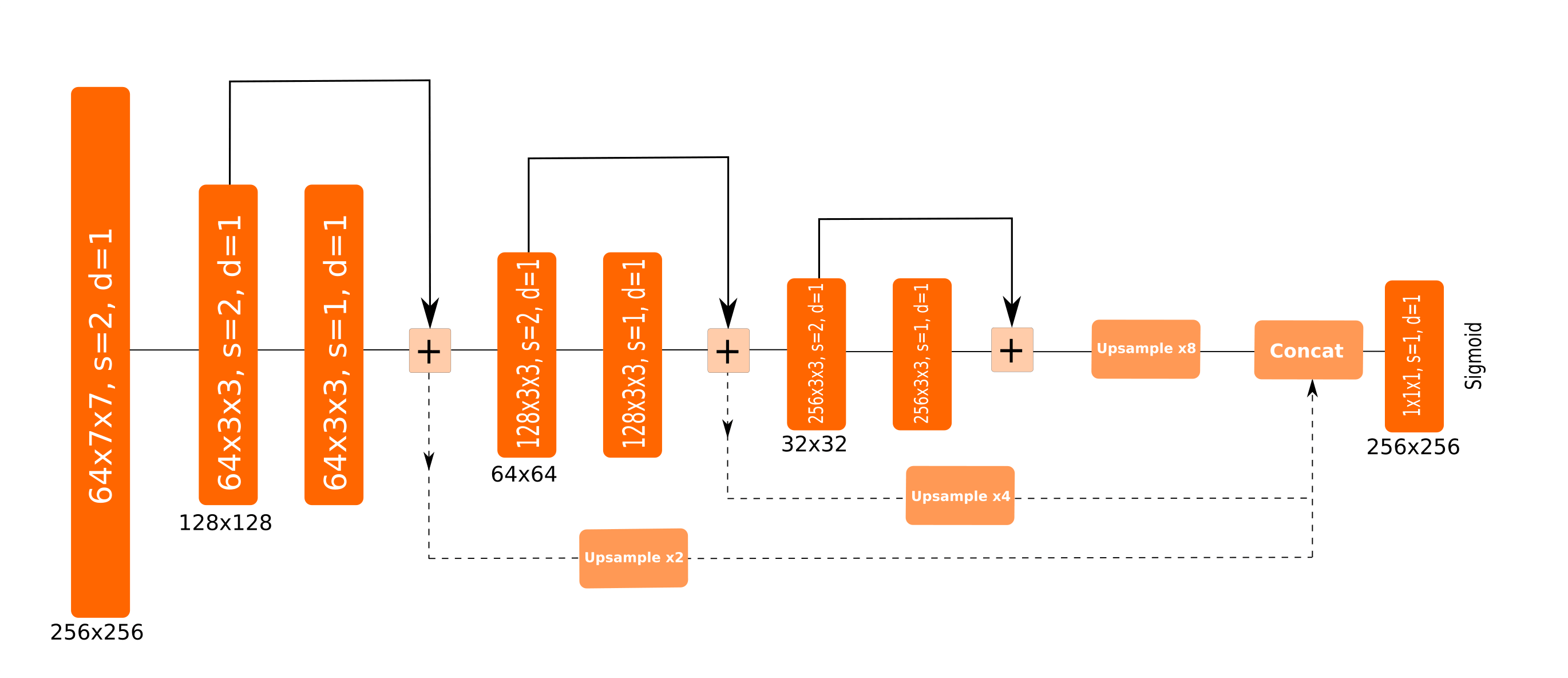}}
\vspace{-0.3cm}
\caption{Detailed overview of SAAD.}
\label{dis}
\end{figure*}

We present additional details about our Generator and Discriminator architectures for reproduction purposes. Fig.\ref{gen} presents the parameters used for each layer of the U-Net Generator while Fig.\ref{dis} shows a detailed overview of SAAD. 

\section{Qualitative results}
\begin{figure*}[h!]
\center
\centerline{\includegraphics[width=17cm]{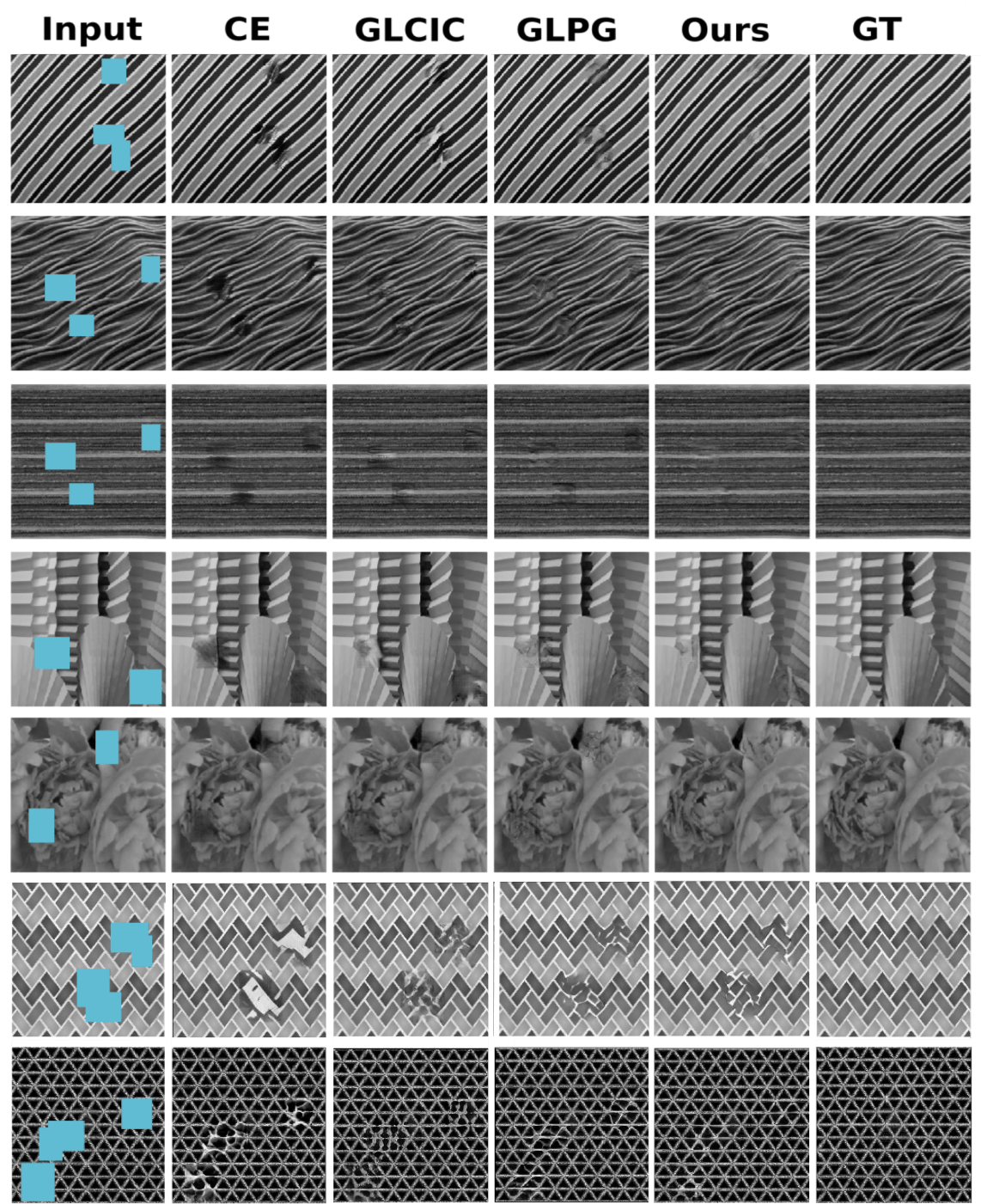}}
\vspace{-0.1cm}
\caption{Qualitative results of different methods for the textures inpainting task (1 of 2).}
\label{dtd}
\end{figure*}

\begin{figure*}[h!]
\center
\centerline{\includegraphics[width=17cm]{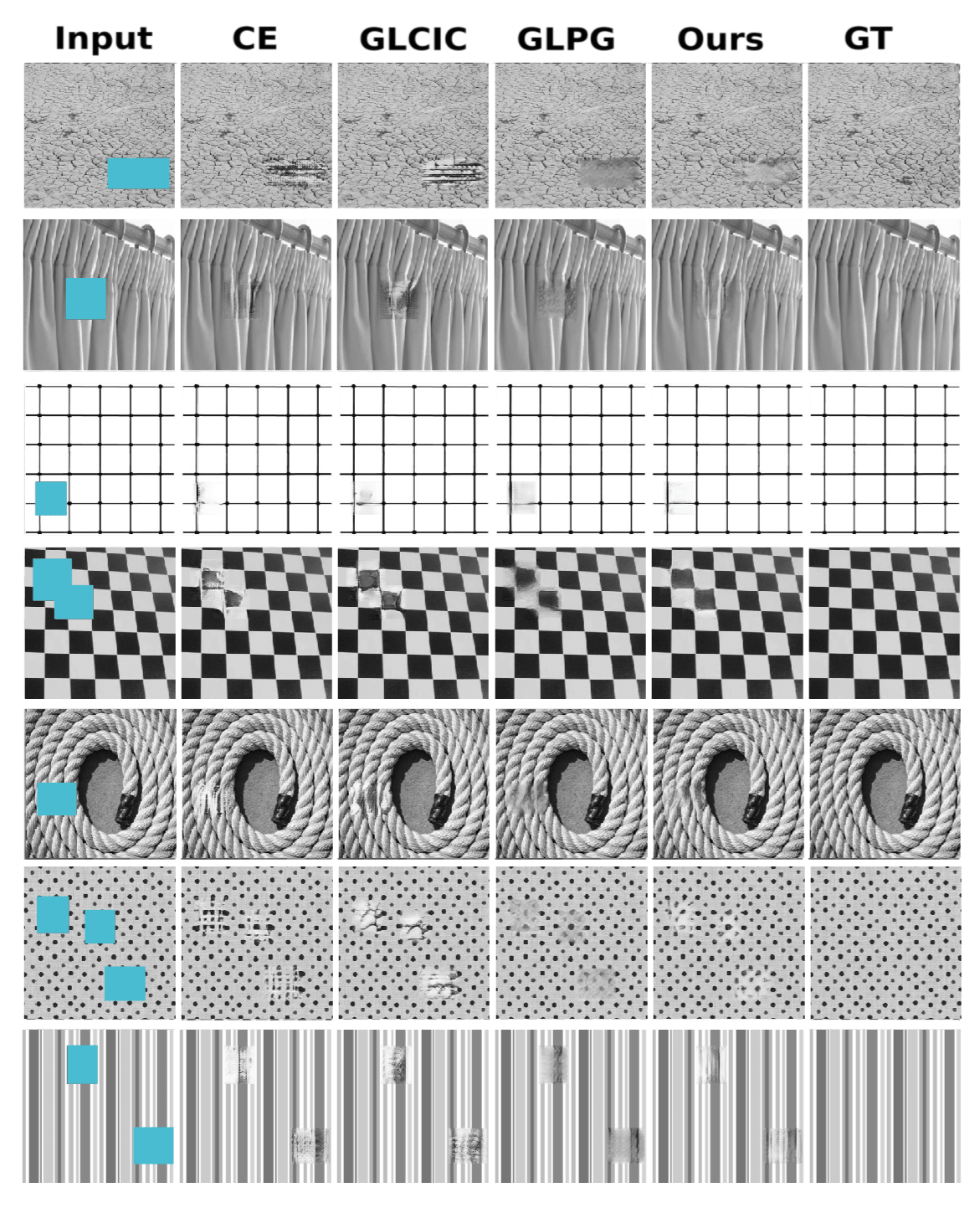}}
\vspace{-0.1cm}
\caption{Qualitative results of different methods for the textures inpainting task (2 of 2).}
\label{dtd_bis}
\end{figure*}

\begin{figure*}[ht!]
\center
\centerline{\includegraphics[width=17cm]{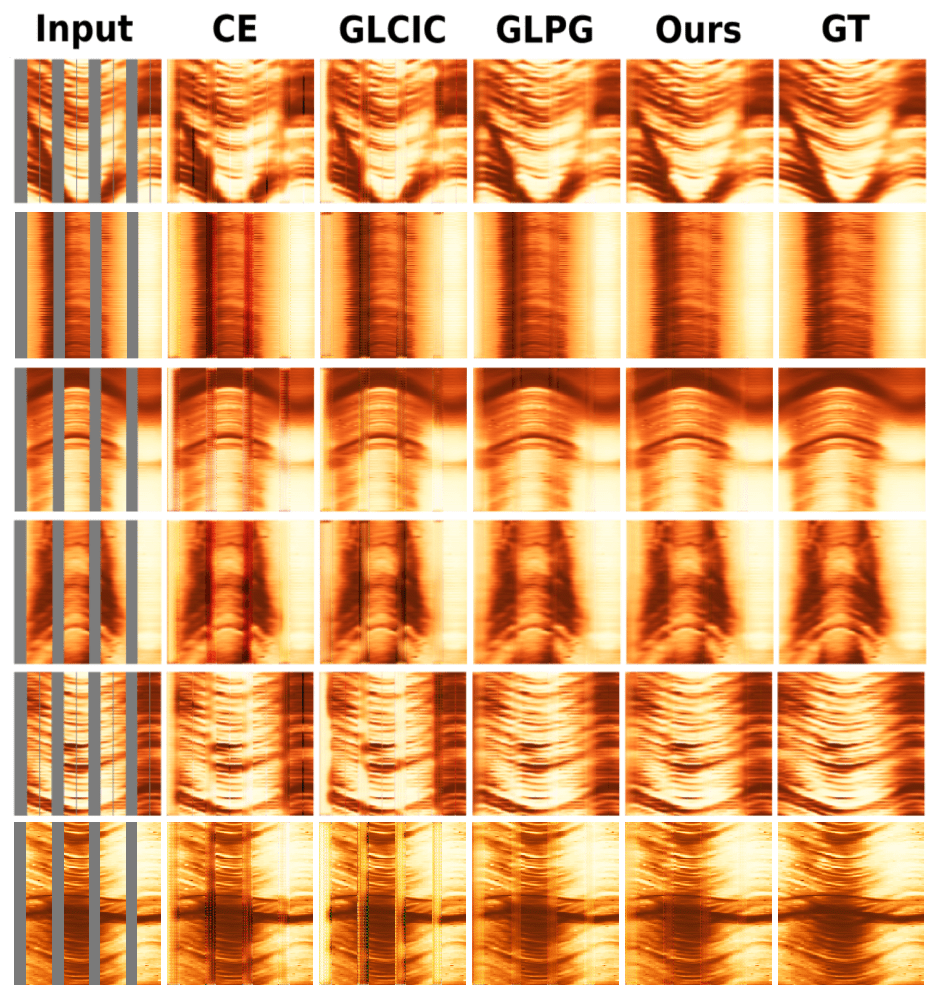}}
\vspace{-0.3cm}
\caption{Qualitative results of different methods for the borehole inpainting task.}
\label{borehole}
\end{figure*}

In this section we present additional qualitative results on both DTD and our borehole dataset. These results show that our proposed method performs better than existing baselines on both datasets. in Fig.\ref{dtd} and Fig.\ref{dtd_bis} we can see that texture are better inpainted (shapes and color continuity are better respected with our method). Wheras in Fig.\ref{borehole} we see that artifacts created by existing inpainting methods are much less visibile when using our approach.

\clearpage

\section{Models performances with respect to coverage for differents methods}
\begin{figure*}[h!]

\centerline{\includegraphics[width=17cm]{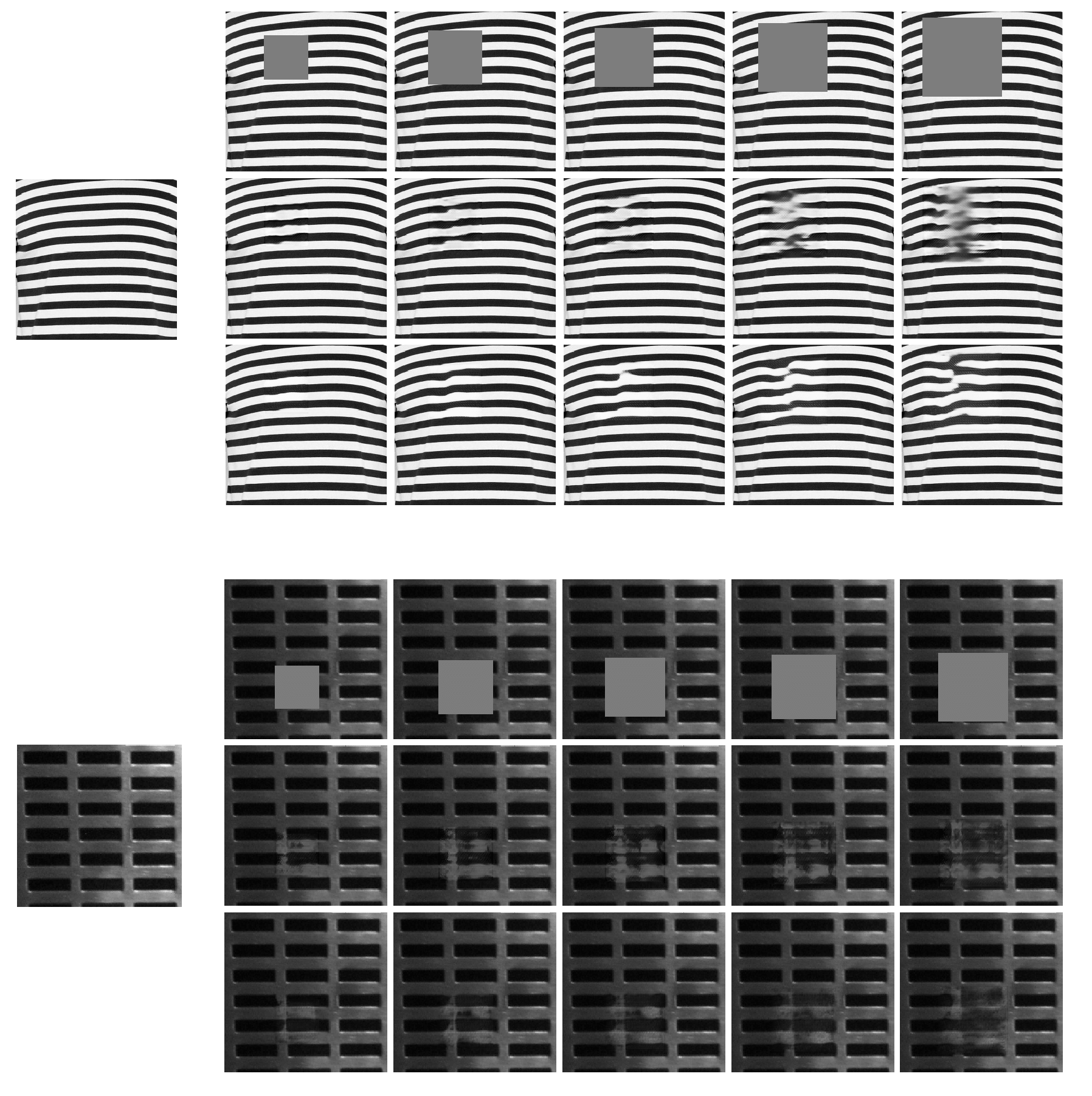}}
\vspace{-0.1cm}
\caption{Inpainting quality with respect to mask size. The leftmost images are the ground-truth. For the rest from top to bottom : Input, output (GLPG), output (Ours).}
\label{size}
\end{figure*}

\begin{figure*}[h!]
\centerline{\includegraphics[width=17cm]{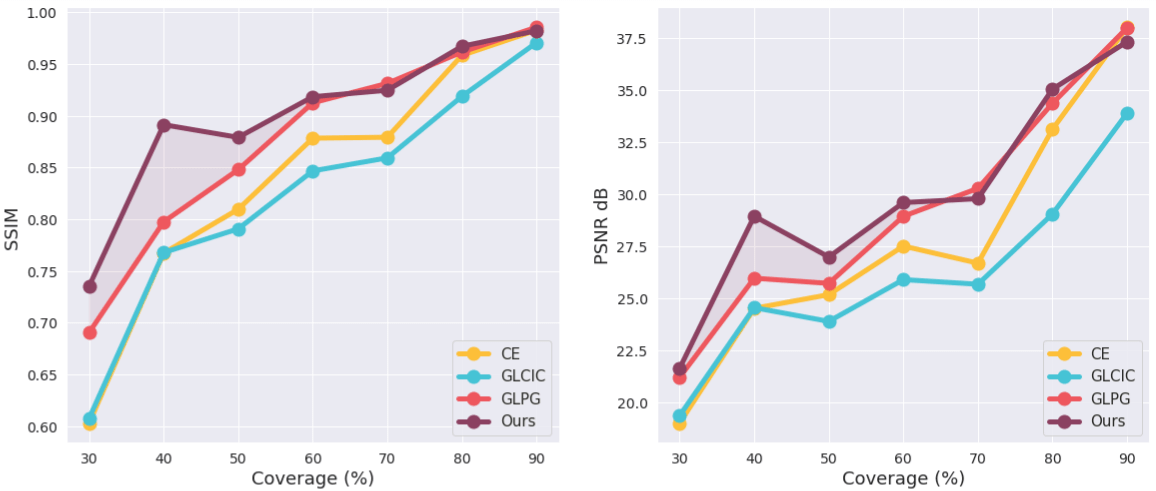}}
\vspace{-0.1cm}
\caption{Comparison of different methods performances in PSNR (dB) and SSIM with respect to coverage(\%)}
\label{qnttve}
\end{figure*}

Fig.\ref{size} shows a comparison between GLPG and our method in terms of inpainting quality w.r.t mask size. The results confirm our intuition that our method is better suited for large zones inpainting that existing methods. These results are confirmed by a quantitative comparison in the main paper and here in Fig.\ref{qnttve}.
